\documentclass{article}

\usepackage{PRIMEarxiv}
\usepackage{verbatim}
\usepackage[utf8]{inputenc} % allow utf-8 input
\usepackage[T1]{fontenc}    % use 8-bit T1 fonts
\usepackage{hyperref}       % hyperlinks
\usepackage{url}            % simple URL typesetting
\usepackage{booktabs}       % professional-quality tables
\usepackage{amsfonts}       % blackboard math symbols
\usepackage{nicefrac}       % compact symbols for 1/2, etc.
\usepackage{microtype}      % microtypography
\usepackage{lipsum}
\usepackage{fancyhdr}       % header
\usepackage{graphicx}       % graphics
\graphicspath{{media/}}     % organize your images and other figures under media/ folder
\usepackage{algorithm}
\usepackage{algpseudocode}
\usepackage{array}
\usepackage[T1]{fontenc}
\usepackage{CJKutf8}
\usepackage[english]{babel}

\usepackage{microtype}
\usepackage{graphicx}
\usepackage{subfigure}
\usepackage{tabularx}
\usepackage{booktabs} % for professional tables

\usepackage{hyperref}

\usepackage{amsmath}
\usepackage{amssymb}
\usepackage{mathtools}
\usepackage{amsthm}
\usepackage{kotex}

\usepackage[capitalize,noabbrev]{cleveref}

\usepackage{enumitem}

\title{Unidirectional Thin Adapter for Efficient Adaptation of Deep Neural Networks
}

\author{
  Han Gyel Sun,  Hyunjae Ahn, HyunGyu Lee, Injung Kim\\
  Handong Global University \\
  \texttt{shggreen@hotmail.com, ahj8762@naver.com,  qwer9497@gmail.com, ijkim@handong.edu} \\
}

\begin{document}
\maketitle

\begin{abstract}
In this paper, we propose a new adapter network for adapting a pre-trained deep neural network to a target domain with minimal computation. The proposed model, unidirectional thin adapter (UDTA), helps the classifier adapt to new data by providing auxiliary features that complement the backbone network. UDTA takes outputs from multiple layers of the backbone as input features but does not transmit any feature to the backbone. As a result, UDTA can learn without computing the gradient of the backbone, which saves computation for training significantly. In addition, since UDTA learns the target task without modifying the backbone, a single backbone can adapt to multiple tasks by learning only UDTAs separately. In experiments on five fine-grained classification datasets consisting of a small number of samples, UDTA significantly reduced computation and training time required for backpropagation while showing comparable or even improved accuracy compared with conventional adapter models.
\end{abstract}

\section{Introduction}
\label{Introduction}
Deep neural networks (DNN) have shown outstanding performance in diverse fields and are developing rapidly. However, DNNs require a large amount of data to learn a target task. There has been a growing body of research that alleviates this issue by adapting a model pre-trained on a large dataset, such as ImageNet, to a small target dataset \cite{simonyan2014very, sun2019meta}. The ‘pre-training followed by fine-tuning’ strategy is effective in improving performance for tasks of insufficient data. In particular, a few recent works propose minimizing or omitting the adaptation step by learning abundant knowledge from a large amount of data with large-scale models such as vision transformer (ViT) and DALL-E \cite{kolesnikov2021image, ramesh2021zeroshot}.\par

However, large-scale DNNs require excessive computational resources for use on low-resource devices and are vulnerable to overfitting when fine-tuned with a small amount of data. Due to these issues, when adapting to new data in an environment with low computing power or insufficient data, many previous works freeze the backbone network and learn only the classifier using transfer learning or few-shot learning \cite{sun2019meta, chobola2021transfer}.\par

Despite the aforementioned issues, the demand for learning or adapting DNNs in the low-resource environment, such as an edge device, is increasing. Training in the client device can improve the accuracy of the target task by further optimizing the pre-trained model with the target data of the user. We can extend this advantage to federated learning that improves a shared model without transferring private data by integrating only the training results of multiple client devices \cite{konevcny2016federated, lim2020federated}. In particular, in certain fields, such as fine-grained classification, adapting the pre-trained feature extractor layers to the target data is crucial because it is difficult to achieve high performance in such fields with only the features learned from a general dataset.\par

To train a DNN in a low-resource environment, it is essential to reduce the computational cost for learning. In previous work, diverse approaches have been proposed to reduce the size and computation of the DNN. Recently, there has been much progress to tackle this issue with lightweight models such as MobileNetV1 and V2 \cite{howard2017mobilenets, sandler2018mobilenetv2}, ShuffleNetV1 and V2  \cite{zhang2018shufflenet, ma2018shufflenet}, EfficientNetV1 and V2 \cite{tan2019efficientnet, tan2021efficientnetv2}. However, even with the lightweight models, training the entire model in a low-resource environment requires additional computation reduction.\par
%Model patch\cite{mudrakarta2018k}, Residual Adapter\cite{rebuffi2017learning}\cite{rebuffi2018efficient}, budget-aware adapter\cite{berriel2019budget}, and FLUTE\cite{triantafillou2021learning} 
\begin{figure*}[t]
   \centerline{\includegraphics[width=0.74\textwidth]{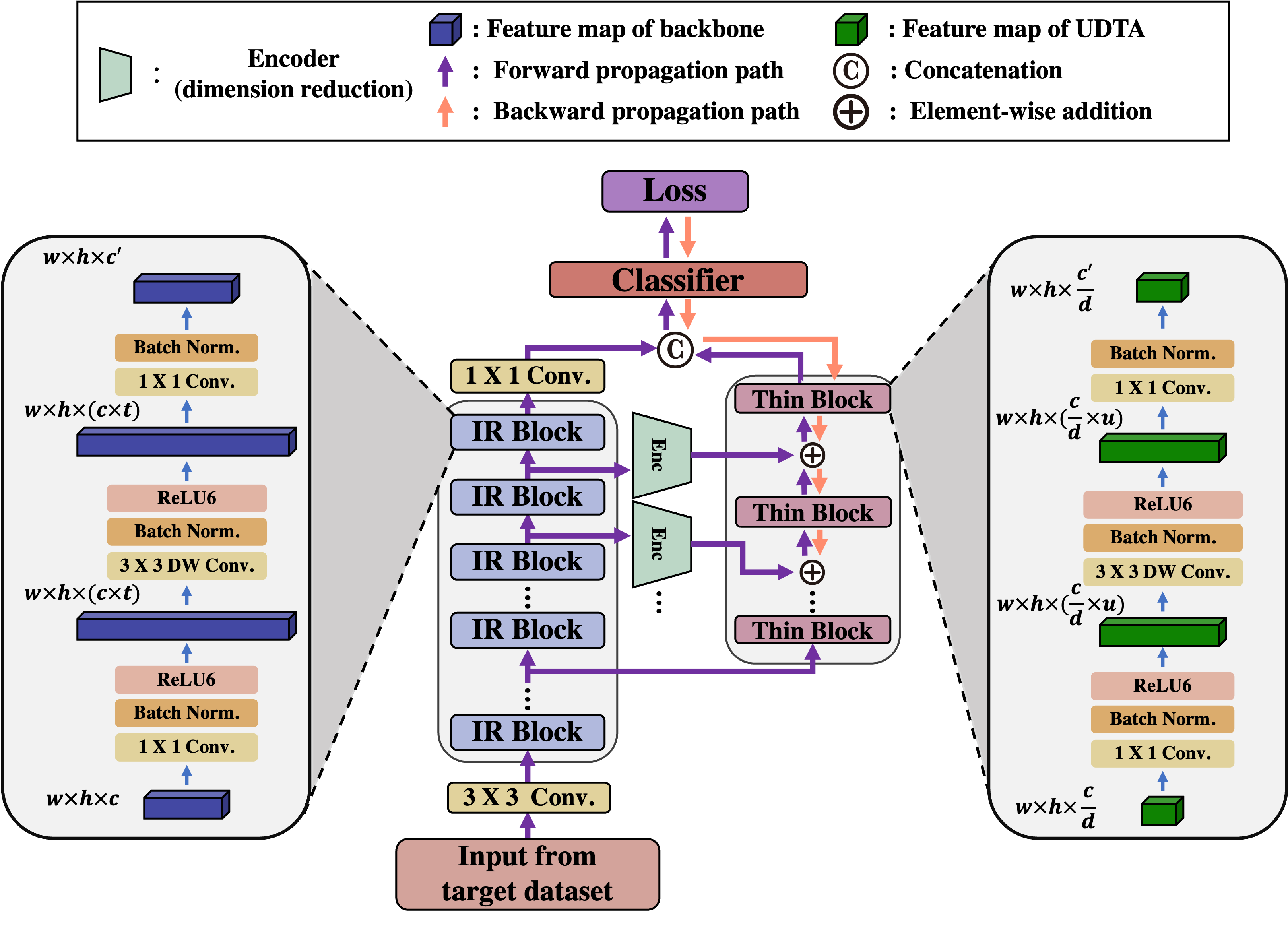}}
\caption{Unidirectional thin adapter (UDTA) combined with a backbone network (MobileNetV2). UDTA adapts the pre-trained backbone network by providing auxiliary features to complement the backbone features. Since the encoders reduce feature dimension, the thin blocks are substantially narrower than IR blocks.}
\label{fig:architecture_of_UDTA}
\end{figure*}
On the other hand, a few previous studies have proposed adapter networks to learn DNNs with only a small amount of data \cite{mudrakarta2018k, rebuffi2017learning, rebuffi2018efficient, berriel2019budget, triantafillou2021learning}.  The adapter improves performance by adjusting the activation of intermediate backbone layers or by providing complement features, without modifying the backbone substantially. Composed of a small number of parameters, the adapters are effective in reducing overfitting and preserving the pre-trained knowledge of the backbone. However, the existing adapters do not reduce computation for training because they still require the gradient of the backbone while training.

To overcome this limitation, in this paper, we propose a unidirectional thin adapter (UDTA) that adapts the pre-trained backbone network to the target task with minimal computation and a small number of target data. UDTA, as shown in \cref{fig:architecture_of_UDTA}, is composed of thin and lightweight blocks that can be trained with minimal computation and a small amount of data. 

Similar to the existing adapters, UDTA utilizes the intermediate features of the backbone to produce features effective for the target task. However, unlike conventional adapters, UDTA has only unidirectional connections from the backbone and does not transmit its features to the backbone. As a result, UDTA does not require the gradient of the backbone during backpropagation, thereby can be trained with minimal computation. In addition, we reduce the dimension of the feature maps delivered from the backbone by autoencoders, further reducing the size and computation of UDTA without significant loss of information. Since UDTA perfectly preserves the pre-trained knowledge of the backbone, a single backbone can conduct multiple tasks by combining a collection of per-task UDTAs. 
In experiments on five fine-grained classification datasets containing a small number of data, UDTA reduced the computation and time for backpropagation by 86.36\% and 77.78\% compared with model patches \cite{mudrakarta2018k}, while exhibiting comparable or improved accuracy compared with the other adapters.

%In experiments on five fine-grained classification datasets containing a small number of data, UDTA reduced the computation and time for training to 46.95\%, 57.14\% compared with residual adapter \cite{rebuffi2017learning, rebuffi2018efficient}, while exhibiting comparable or substantially improved absolute accuracy compared with the other adapters. In particular, the calculation and time required for backpropagation are reduced by 85.14\% and 75.40\%, respectively, compared with residual adapter.

Our main contributions include 1) a theoretical analysis of the reason why the existing adapter networks fail to reduce computation for training, 2) a novel unidirectional thin adapter (UDTA) to adapt a pre-trained DNN to a new dataset with minimal computation and data, 3) a lightweight adapter structure and reduced data transfer due to dimensionality reduction, and 4) comparable or improved accuracy compared with existing adapter networks for five small-size datasets for fine-grained classification.

Section 2 of this paper introduces related work. Section 3 analyzes the problems with conventional adapter networks in terms of computational efficiency in backpropagation and proposes the unidirectional thin adapter to overcome the limitation. Section 4 and Section 5 present experimental results and conclusions, respectively.

\section{Related Work}
\label{Related Work}

\begin{figure}[t] 
   \centerline{\includegraphics[width=0.5\linewidth]{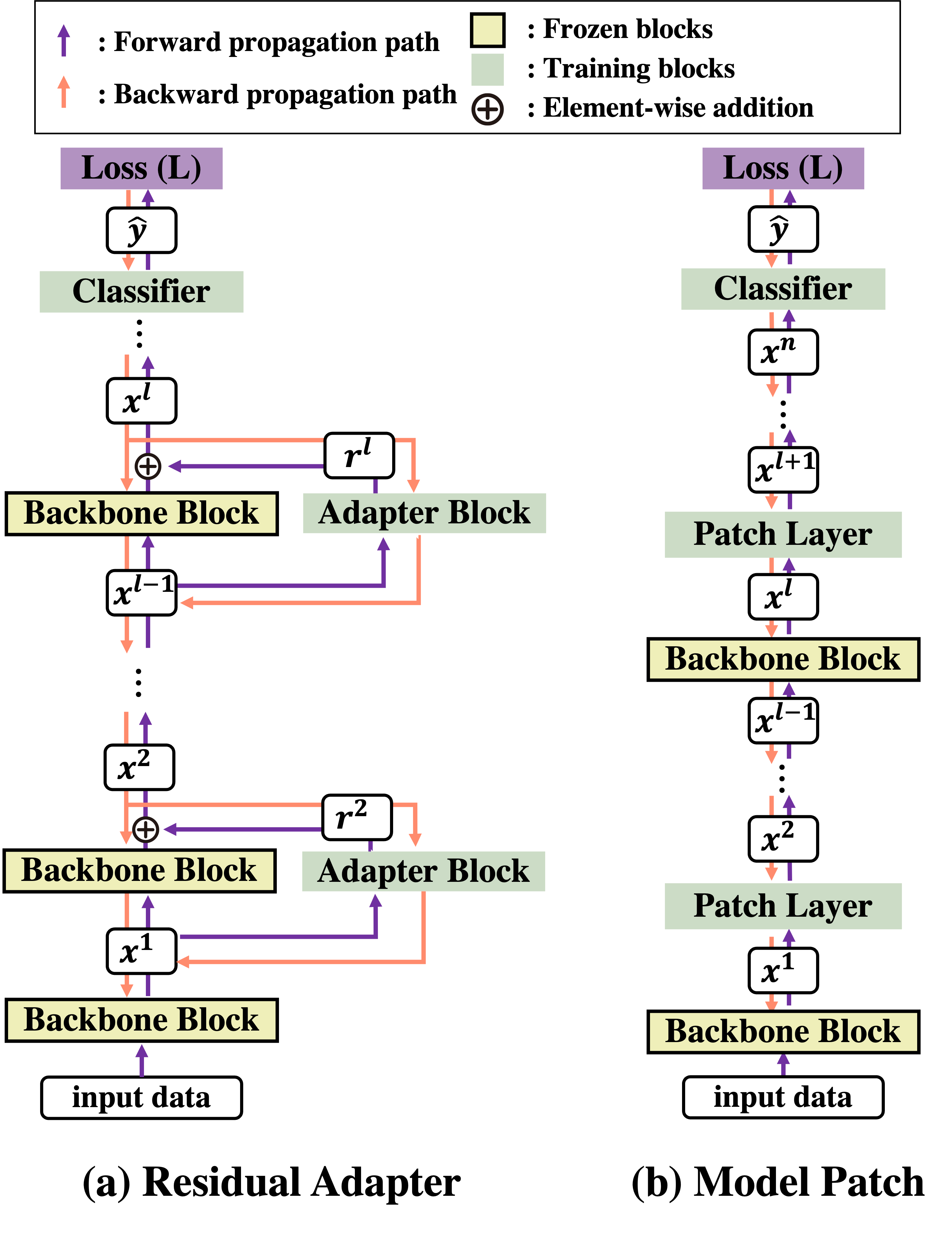}}
\caption{Training of residual adapters and model patches. The red arrows illustrate the backpropagation flow. Both adapters require the gradient of the backbone blocks.}
\label{fig:conventional_backprop}
\end{figure}

\textbf{Pre-training followed by fine-tuning strategy} In recent years, there has been an increasing interest in domain adaptation by fine-tuning a pre-trained DNN model. Specifically, the `pre-training followed by fine-tuning' strategy, that first pre-trains a DNN with a large volume of data, such as ImageNet, and then fine-tunes it with a small amount of target data, has been widely adopted and demonstrated its effectiveness in improving performance for the tasks with insufficient training data. \cite{sun2019meta} proposes a meta transfer learning method for effectively fine-tuning a DNN pre-trained on a large dataset for multiple few-shot tasks. \cite{chobola2021transfer} applies a pre-trained convolutional neural network (CNN) to map samples to a latent space, and then classifies the latent feature using a simple model, such as k-NN classifier and Gaussian mixture models.\par

\textbf{Lightweight network models} MobileNetV2 \cite{sandler2018mobilenetv2}, ShuffleNetV2 \cite{ma2018shufflenet}, and EfficientNetV2 \cite{tan2021efficientnetv2} propose lightweight DNN models with reduced size and computation. TinyTL \cite{cai2020tinytl} proposes a memory-efficient model that learns only the bias without storing intermediate activation. Besides the above models, researchers have attempted to reduce the computation or the number of parameters of the DNN. Lightlayers \cite{jha2020lightlayers} and batch ensemble \cite{wen2020batchensemble} propose network modules as to reduce computation. Lightlayers decompose convolutional layers and fully connected layers into a combination of minor operations. In contrast, batch ensemble approximates a weight matrix by the product of a single row and a single column. % The autoencoder\cite{hinton2006reducing} focuses on compressing feature minimizing the loss of information by learning a symmetrical neural network structure with the same input and output values.

We built our backbone network based on MobileNetV2 \cite{sandler2018mobilenetv2}. The main building block of MobileNetV2 is the inverted residual (IR) block that was designed to minimize computational requirement and the number of parameters preserving performance. The IR block is composed of a 1x1 convolution that increases the number of channels, a 3x3 depth-wise convolution, and another 1x1 convolution that decreases the number of channels. The third 1x1 convolution layer omits ReLU activation to reduce information loss.\par
\begin{figure}[h] 
   \centerline{\includegraphics[width=0.4\linewidth]{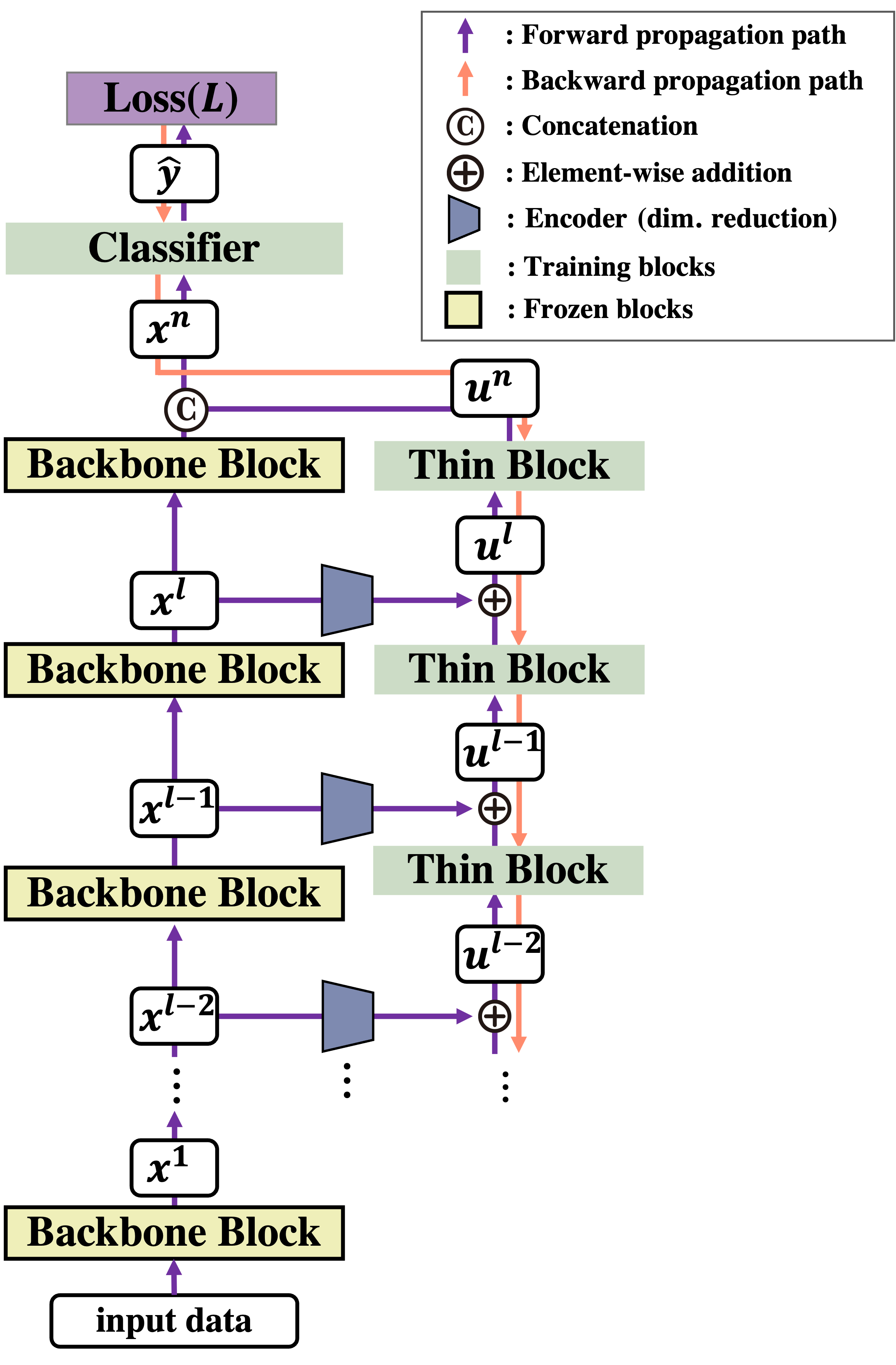}}
\caption{Training of UDTA. The red arrows illustrate the backpropagation flow. UDTA learns without the gradient of the backbone.}
\label{fig:backprop_UDTA}
\end{figure}
\textbf{Adapter networks} Numerous previous studies on domain adaptive learning have proposed adapter networks to adapt a pre-trained DNN to a new target domain efficiently \cite{farahani2021brief, mudrakarta2018k, rebuffi2017learning, rebuffi2018efficient, berriel2019budget, triantafillou2021learning}. Composed of a small number of parameters, the adapter networks can be trained with a small number of target data.

%For reducing the parameters in domain adaptive learning \cite{farahani2021brief}, model patch  \cite{mudrakarta2018k}, residual adapter \cite{rebuffi2017learning}\cite{rebuffi2018efficient}, budget-aware adapter \cite{berriel2019budget}, and FLUTE \cite{triantafillou2021learning} propose an efficient learning method for the target domain using a small number of parameters and a small dataset.\par

Some previous works have applied partial learning to reduce the number of parameters \cite{farahani2021brief, mudrakarta2018k}. \cite{mudrakarta2018k} proposes to adapt a pre-trained DNN to a new task by learning only a small number of layers, fixing the other parts. They demonstrate that training only a few layers with a small number of parameters, such as scale-and-bias patch and depthwise-convolution patch, can significantly improve performance for a new target task. \cite{rebuffi2017learning, rebuffi2018efficient} proposes a residual adapter that reduces the number of parameters to train in domain adaptation. While \cite{mudrakarta2018k} reuses the layers of the pre-trained network for model patches, \cite{rebuffi2017learning, rebuffi2018efficient} adapt the backbone by combining it with a collection of adapter networks and learning only the adapters. The adapter network is composed of layers with a small number of parameters, such as 1x1 convolution and batch normalization. They combine residual adapters to all 3x3 convolution layers of the backbone network. Model patches and residual adapters help DNNs adapt to a new domain with a small number of training samples. However, they do not reduce the computational cost for training, as analyzed in the next section.

%In model patches, partial learning is derived as a method of partially learning a model by configuring a patch of layers with batch regularization or depth-wise convolution operation that consists of few parameters. Using the knowledge of the pre-trained model, model patch showed that adaptive learning is possible for the user’s target domain with only a few parameters.

%Subsequently, residual adapter \cite{rebuffi2017learning, rebuffi2018efficient} using the residual block \cite{li2021universal, he2016deep} also reduced the number of parameters used in the domain adaptation process. Residual Adapter uses a method different from the model patch method, which reuses parameters belonging to the pre-trained model for domain adaptive learning. Residual Adapter is a set of 1×1convolutional layers composed of a few parameters and is attached to each 3×3 layer of the model in series or parallel to be used for domain adaptive learning. The residual adapter enables effective domain adaptive learning on various datasets, even with a few parameters, and the performance does not significantly decrease compared to the models trained on each dataset.\par

\cite{berriel2019budget,triantafillou2021learning} propose a budget-aware adapter ($BA^2$) and FLUTE to reduce the computational cost. Budget-aware adapter selects the most relevant feature channels and saves the computational cost by activating only the selected channels. \cite{triantafillou2021learning} proposes a method for few-shot learning with a universal template (FLUTE), a partial model that can define a wide array of dataset-specialized models by plugging in appropriate components. They are effective in reducing computation for forward propagation, but cannot be used to reduce computation for training.

\section{Efficient Domain Adaptation with Unidirectional Thin Adapter}

\subsection{Backpropagation of Conventional Adapter Networks}
\begin{figure*}[t]
   \centerline{\includegraphics[width=0.79\textwidth]{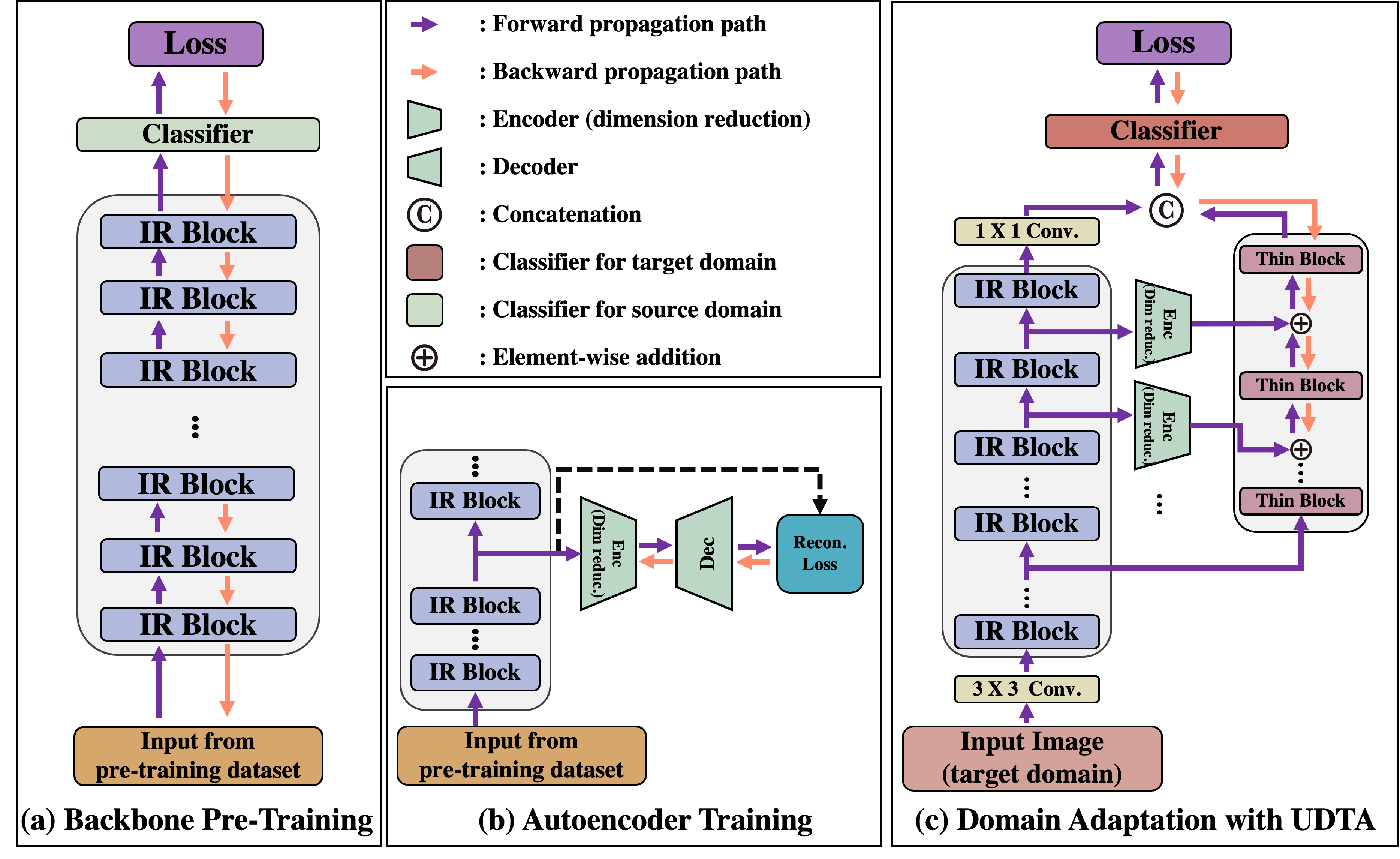}}
\caption{Training of UDTA. (a) and (b) are conducted on a server, while (c) is conducted on the client device.}
\label{fig:train_process_of_UDTA}
\end{figure*}
The gradient descent algorithm optimizes the neural network by adjusting model parameters to minimize loss function as Eq. (\ref{eq:gradient_descent_optimizer}), where $w^l$, $L$, and $\eta$ denote the weights of the $l$-th layer, loss function, and learning rate, respectively.

\begin{equation}
    w^l \leftarrow  w^l -\eta \frac{\partial L}{\partial w^l} 
\label{eq:gradient_descent_optimizer}
\end{equation}

As shown in Eq. (\ref{eq:gradient_descent_optimizer}), the gradient descent algorithm requires the gradient $\frac{\partial L}{\partial w^l}$. The backpropagation algorithm computes the gradient sequentially from the top to the bottom layer.

Residual adapters are combined with the backbone layers in a parallel or serial way. \cref{fig:conventional_backprop}(a) displays parallel residual adapters attached to the backbone network. In training procedure, the backbone blocks are fixed and only the adapter blocks are trained with the target data. 
To train the weights of the residual adapter $w_r^l$, we need to compute its gradient $\frac{\partial L}{\partial w_r^l}$. 

The red arrows in \cref{fig:conventional_backprop}(a) illustrate the backpropagation path to learn the residual adapters. Applying the chain rule, $\frac{\partial L}{\partial w_r^l}$ is decomposed as Eq. (\ref{eq:gradient_of_residual_adapter}), where $x^l$ and $r^l$ denote activation of the $l$-th block in the backbone and the residual adapter, respectively.\par

\begin{equation}
    \frac{\partial L}{\partial w_r^l} = \frac{\partial L}{\partial r^l} \frac{\partial r^l}{\partial w_r^l} = \frac{\partial L}{\partial x^l} \frac{\partial x^l}{\partial r^l} \frac{\partial r^l}{\partial w_r^l}
\label{eq:gradient_of_residual_adapter}
\end{equation}

$\frac{\partial L}{\partial x^l}$ in Eq. (\ref{eq:gradient_of_residual_adapter}) is computed by combining the gradient from $x^{l+1}$ and that from $r^{l+1}$ as Eq. (\ref{eq:backbone_gradient_of_residual_adapter}). The recursive expansion of Eq. (\ref{eq:backbone_gradient_of_residual_adapter}) includes $\frac{\partial L}{\partial x^{n}} (\prod_{k=l}^{n-1} \frac{\partial x^{k+1}}{\partial x^k})$.\par

\begin{equation}
     \frac{\partial L}{\partial x^{l}} = \frac{\partial L}{\partial x^{l+1}}\frac{\partial x^{l+1}}{\partial x^{l}} + \frac{\partial L}{\partial r^{l+1}}\frac{\partial r^{l+1}}{\partial x^{l}}
\label{eq:backbone_gradient_of_residual_adapter}
\end{equation}

As shown in Eq.(\ref{eq:gradient_of_residual_adapter}) and (\ref{eq:backbone_gradient_of_residual_adapter}), computing $\frac{\partial L}{\partial w_r^l}$ requires the gradients $\frac{\partial L}{\partial x^k}$ for all $k \geq l$. Consequently, training the parallel residual adapter requires backpropagation through the entire network including the backbone. The computation of the gradients to learn serial adapters is similar to that with model patches explained below.

\cref{fig:conventional_backprop}(b) illustrates the forward and backward propagation of a DNN with model patches. The gradient for the weight $w_p^l$ of the $l$-th patch layer is computed as Eq.(\ref{eq:gradient_of_model_patch}), where $x^l$ is the activation of a patch layer and $x^k$s are the activation of either a patch layer or a backbone layer.\par

\begin{equation}
    \frac{\partial L}{\partial w_p^l} = \frac{\partial L}{\partial x^l}\frac{\partial x^l}{\partial w_p^l} = \frac{\partial L}{\partial x^{n}} 
    (\prod_{k=l}^{n-1} \frac{\partial x^{k+1}}{\partial x^k})
    \frac{\partial x^l}{\partial w_p^l}
    \label{eq:gradient_of_model_patch}
\end{equation}

Eq. (\ref{eq:gradient_of_model_patch}) shows that the training of model patches also requires backpropagation through the entire network.

\subsection{Structure of Unidirectional Thin Adapter}

\cref{fig:architecture_of_UDTA} illustrates the structure of UDTA combined with a backbone network. In \cref{fig:architecture_of_UDTA}, $w$, $h$, and $c$ denote the width, height, and the number of channels of the backbone feature map, respectively. $d > 1$, $t > 1$ and $u > 1$ denote the dimensionality reduction factor of the encoder, channel expansion factor of the IR block, and channel expansion factor of thin block, respectively.\par

%the structure consists of the backbone network, the encoder, and UDTA. The backbone model extracts high-level features based on the pretrained knowledge of the source dataset. The encoder receives the extracted features from the layers of the backbone network and reduce the dimension, minimizing the loss of information. We add the features with the reduced dimension from the encoder and the output feature from the previous thin block to be the input for the next thin block so that it efficiently preserves the knowledge of the backbone network. The UDTA learns the domain specific knowledge with the preserved knowledge of the backbone network. Finally, we concatenate the output feature from the backbone and the UDTA, which conclusively become the input of the classifier for predicting each class. The classifier predicts each class of data by utilizing the concatenated output of the resulting feature from UDTA and the backbone network.

The backbone network is pre-trained with a large-scale dataset and extracts high-level features from the input. The UDTA is composed of a stack of thin blocks. The dimension of the UDTA feature map is lower than that of the backbone feature map at the same level. Each thin block takes as input the feature map of the corresponding backbone block, and propagates features to upper blocks to produce auxiliary features for the classifier. In training, we freeze the backbone and train only UDTA with the target data to produce auxiliary features that complement the backbone features, thereby improving the performance for the target task. In other words, the backbone learns general knowledge from a large-scale pre-training dataset, while UDTA learns domain-specific knowledge from the target data to complement the backbone features. The classifier combines the features from the backbone and UDTA and outputs the classification result.

The encoders between the backbone and the UDTA blocks reduce the dimension of the features transmitted from the backbone, which allows further reducing the size and computation of UDTA. We use autoencoders to train UDTA encoders to reduce feature dimension while minimizing information loss. As a result, the UDTA comprises a small number of parameters and requires a small amount of computation for both forward and backward propagation.

% Storing pretrained parameters of the entire model of the multiple domains of  domain adaptation requires an inefficiently significant amount of time and resources. Training only the classifier using transfer learning has a significant drawback, such as low recognition performance. To address these issues, we propose an adapter structure that can be adapted to each domain. The existing adapters receive output features from each layer of the backbone network to effectively utilize the knowledge of the backbone network. However, because the existing adapter structures require an extensive  amount of computation during the backpropagation process, we designed the adapter as a unidirectional structure. The unidirectional adapter reduces the computation for backpropagation by eliminating the gradient flow towards the backbone from the adapter. Additionally, the adapter preserves the knowledge of the backbone network with a reduced feature dimension by utilizing the encoder. The encoder diminishes the width of the adapter, which in turn reduces the computation and number of parameters for the entire model. The detailed structure of the model to which UDTA was applied is shown in \cref{fig:architecture_of_UDTA}. 

The full training of the entire network requires a large amount of computation. On the other hand, freezing the backbone and training only the classifier reduce computation significantly but exhibit limited performance because of the lack of domain-specific knowledge. UDTA can make a good trade-off between performance and computational cost. In addition, UDTA perfectly preserves the pre-training knowledge of the backbone since it learns without modifying the backbone. Therefore, it is possible to perform multiple tasks with a single backbone network by adapting only UDTAs to the target tasks. This is an important advantage, especially in the low-resource environment.

The most prominent advantage of UDTA over existing adapter networks is that we can train it with a minimal amount of computation. In Section 3.1, we have shown that residual adapters and model patches cannot reduce computation for training, because training them requires the gradient of the backbone which demands heavy computation. To learn without the gradient of the backbone, we designed UDTA to be connected to the backbone in only one direction. While it receives the intermediate features from multiple backbone blocks, it does not provide any feature to the backbone blocks.

Training UDTA requires the gradient $\frac{\partial L}{\partial w_{u}^{l}}$, where $w_u^l$ denotes the weights of the $l$-th thin block. The gradient is computed as Eq. (\ref{eq:backpropagation_UDTA}).\par

\begin{equation}
    \begin{aligned}
        \frac{\partial L}{\partial w_{u}^{l}} &= \frac{\partial L}{\partial u^{l}}\frac{\partial u^{l}}{\partial w_{u}^{l}}
       = \frac{\partial L}{\partial u^{n}}\frac{\partial u^{n}}{\partial u^{l}} \frac{\partial u^{l}}{\partial w_{u}^{l}}  \\
       &= \frac{\partial L}{\partial u^{n}} (\prod_{k=l}^{n-1} \frac{\partial u^{k+1}}{\partial u^k}) \frac{\partial u^l}{\partial w_u^l}
    \end{aligned}
    \label{eq:backpropagation_UDTA}
\end{equation}

Note that Eq. (\ref{eq:backpropagation_UDTA}) does not require the gradient of the backbone $\frac{\partial L}{\partial x^l}$ for any $l$. \cref{fig:backprop_UDTA} illustrates the backpropagation path to compute gradient $\frac{\partial L}{\partial w_{u}^{l}}$ to train UDTA. The red arrow starts from the loss and goes through the thin blocks. However, it does not pass any backbone block. Consequently, UDTA can learn without backpropagation through the entire backbone network, and therefore, can save computation for training substantially.

%The adapter networks proposed in 3.1 have a drawback in that it is difficult to learn the adapter without the gradient of the backbone during the domain adaptation learning process. To address this issue, the connection between the adapter and backbone can be eliminated. To implement this approach, we propose a new adapter connection structure, as shown in \cref{fig:backprop_UDTA}.\par
%As the first thin block shown in \cref{fig:architecture_of_UDTA}, the output feature of the IR block becomes the direct input of the thin block when the corresponding backbone layer has a low output dimension. However, when it is connected to the backbone layer that outputs a high-dimensional feature map, the amount of computation and the number of learning parameters are reduced by reducing the dimension through the encoder. The encoder used was trained based on the autoencoder structure for the dataset used in the pretraining. Additionally, the final output value of the backbone network and the final output value of UDTA were concatenated and input to the classifier layer. \par

The structure of UDTA is based on the inverted residual block \cite{sandler2018mobilenetv2}, as shown in \cref{fig:architecture_of_UDTA}. The first $1 \times 1$ convolution increases the number of channels, and the $3 \times 3$ depth-wise convolution extracts spatial information from the input feature. The last 1 × 1 convolution decreases the number of channels. The width and depth of UDTA can be flexibly decided according to the characteristics of the target data and the backbone network. The autoencoder combines an encoder and a decoder, each of which is composed of a $1\times1$ convolution layer. The input dimension of the autoencoder is the same as the output dimension of the backbone layer, while its hidden dimension is the same as the input dimension of the UDTA layer.

%Such an inverted residual block-based structure exhibits descent  relatively fewer parameters than general residual blocks and designing a lightweight deep neural network model suitable for an end device. As well, there is an advantage that the width and depth of each layer can be flexibly changed and configured according to the characteristics of the model or domain. \par
% UDTA is calculated similarly to the parallel adapter proposed in the residual adapter. The network with UDTA has a structure in which an input connection from the backbone network to the adapter exists during the learning process, whereas an output  connection to the central neural network does not exist.
% \par
% Similarly, $\frac{\partial L}{\partial u^{l}}$ requires gradients for subsequent thin blocks as $\frac{\partial L}{\partial u^n}$ and $\frac{\partial u^{k+1}}{\partial u^k}$ and does not require the gradient of the backbone neural network. Therefore, in the model to which UDTA is applied, backpropagation proceeds in a direction that does not calculate the gradient $\frac{\partial L}{\partial x^l}$ for the backbone network. In this structure, backpropagation is performed only on a thin block composed of a relatively small scale in the domain adaptive learning process, and it is possible to significantly reduce the amount of computation compared to the existing methods.

\subsection{Learning of Recognition System Using Unidirectional Thin Adapter}
We train the entire network in three steps.
% \begin{enumerate}
% 	\item  Pre-train the backbone network with a large-scale dataset on a server.
% 	\item  Train the encoders for dimensionality reduction using autoencoders on a server. With the pretrained backbone network, we train autoencoders to reproduce the activation of the intermediate backbone blocks through hidden feature vectors with a lower dimension. Then, separate the encoders of the trained autoencoders from the decoder, and attach it to the backbone for dimensionality reduction. 
% 	\item Combine UDTA, the pretrained backbone and encoders, and the classifier, freeze the pre-trained encoder and backbone, and learn only the UDTA and classifier using the target data on the client device.
% \end{enumerate}
In the first step, the backbone network learns abundant general knowledge that is useful for many target tasks, from a large-scale dataset such as ImageNet on a server, as shown in \cref{fig:train_process_of_UDTA}(a). 

\begin{figure}[t]
   \centerline{\includegraphics[width=0.7\linewidth]{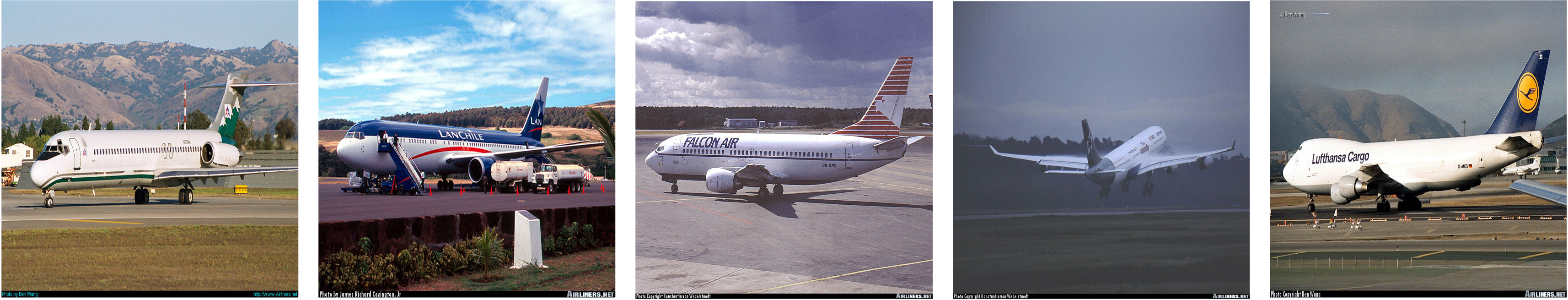}}
\caption{FGVC-Aircraft image samples. Only minor differences distinguish the aircraft types.}
\label{fig:aircraft_sample}
\end{figure}
\begin{table}[t]

\centering
\caption{The number of classes and samples in the datasets for fine-grained classification. (Air: FGVC-Aircraft, Cars: Stanford Cars, DTD: Describable Textures, Dogs: Stanford Dogs, Flow: Flowers-102)}
\resizebox{0.6\linewidth}{!}{%

\begin{tabular}{|c|c|c|c|c|}
\hline
     & Classes & Training Samples             & Test Samples & Total Samples \\ \hline
Air  & 100     & 6,667                        & 3,333        & 10,000        \\ \hline
Cars & 196     & 8,144                        & 8,041        & 16,185        \\ \hline
DTD  & 47      & 3,760                        & 1,880        & 5,640         \\ \hline
Dogs & 102     & 2,040                        & 6,149        & 8,189         \\ \hline
Flow & 120     & 12,000                       & 8,580        & 20,580        \\ \hline
\end{tabular}%
}

\label{tab:num_data}
\end{table}

\begin{table}[!h]
\caption{Experimental settings for backbone pre-training, autoencoder training, and domain adaptation}
\centering
\resizebox{0.6\linewidth}{!}{%
\begin{tabular}{|c|c|c|c|c|c|}
\hline
 & Epoch & Learning rate & Batch & Optimizer                                                        & Loss function \\ \hline
Pre-training  & 120   & 0.01          & 256   & \begin{tabular}[c]{@{}c@{}}ADAM\\ (a=0.9, b= 0.999)\end{tabular} & Cross Entropy \\ \hline
AE training   & 120   & 0.001         & 256   & \begin{tabular}[c]{@{}c@{}}ADAM\\ (a=0.9, b= 0.999)\end{tabular} & MSE           \\ \hline
Domain adap.   & 120   & 0.001         & 256   & \begin{tabular}[c]{@{}c@{}}ADAM\\ (a=0.9, b= 0.999)\end{tabular} & Cross Entropy \\ \hline
\end{tabular}%
}

\label{tab:Experimental Settings}
\end{table}

The second step is to learn the encoders for dimensionality reduction. Similar to the first step, we also train the autoencoder with a large-scale pre-training dataset on a server, as shown in \cref{fig:train_process_of_UDTA}(b). To reduce feature dimension while minimizing the loss of information, we train the autoencoder by the $L_2$ reconstruction loss. 
We do not train the backbone and the autoencoders at the same time because, before the backbone training is complete, learning to compress the immature features is meaningless. After the second step, we use the encoder for dimensionality reduction and discard the decoder.

In the final step, we freeze the backbone and autoencoder, and train only UDTA and the classifier with the target dataset. Due to the encoder, the feature map of UDTA has significantly lower dimension compared with that of the backbone, and therefore, UDTA can learn from a small amount of data with a minimal amount of computation. Furthermore, we combine only a few top layers with thin layers, further saving parameters and computation.
However, for the backbone block with low dimensionality, we directly connect the thin block to the backbone block without encoder.

\section{Experiment}
\subsection{Datasets and Experimental Settings}
\textbf{Datasets}
We used the ImageNet dataset \cite{russakovsky2015imagenet} for the pre-training of the backbone network and the autoencoders. For the target data, we used five small-size datasets designed for fine-grained classification: FGVC-Aircraft \cite{maji2013fine}, Stanford Cars \cite{krause20133d}, DTD \cite{cimpoi2014describing}, Oxford 102 Flowers \cite{nilsback2008automated}, and Stanford Dogs \cite{khosla2011novel} datasets. They consist of a small number of samples per class, and the difference between class images in those datasets is small compared with ordinary datasets. \cref{fig:aircraft_sample} displays a few samples in FGVC-Aircraft set.  To obtain high performance for those datasets, it is essential to learn discriminative features by domain adaptation. The number of classes and samples are presented in \cref{tab:num_data}.

%FGVC-Aircraft \cite{maji2013fine} dataset consisting of aircraft images for the domain adaptive learning. It consists of 100 classes, 67 training samples and 33 test samples for each class.

%Besides FGVC-Aircraft, Stanford Cars \cite{krause20133d}, CUB-200-2011 \cite{wah2011caltech}, DTD \cite{cimpoi2014describing}, Stanford Dogs \cite{khosla2011novel},and Oxford 102 Flowers \cite{nilsback2008automated} were used to evaluate the performance for adapting to different domains. 
%Because these datasets are organized within a limited category, they have unique characteristics that distinguish them from each other. These characteristics reflect the particular case of training on a user’s end device with a domain that is composed of a diverse and small amount of dataset depending on the circumstances of the user’s end-devices.

\begin{table}[t]
\caption{Computation, time, and accuracy of UDTA compared with other models. (Scratch: backbone trained from scratch, Full-FT: fully fine-tuned backbone, Top-FT: backbone with fine-tuned top layer, MP: model patch \cite{mudrakarta2018k}, RA: residual adapters \cite{rebuffi2017learning,rebuffi2018efficient}, and UDTA: proposed method)}
\centering
\resizebox{0.6\linewidth}{!}{%
\begin{tabular}{|c|r|r|r|r|r|r|}
\hline
\multicolumn{1}{|l|}{} & \multicolumn{1}{c|}{F.FLOPs} & \multicolumn{1}{c|}{B.FLOPs} & \multicolumn{1}{c|}{F.Time} & \multicolumn{1}{c|}{B.Time} & \multicolumn{1}{c|}{Acc.} & \multicolumn{1}{c|}{Params} \\ \hline
Scratch                     & 85,322M                      & 170,645M                     & 8.2ms                      & 12.6ms                      & 63.57                    & 2352K                       \\ \hline
Full-FT                       & 85,322M                      & 170,645M                     & 8.2ms                      & 12.6ms                      & 81.03                     & 2352K                       \\ \hline
Top-FT                        & 85,322M                      & 66M                          & 7.8ms                      & 0.4ms                      & 41.91                     & 129K                        \\ \hline
MP                            & 85,322M                      & 165,096M                     & 8.2ms                      & 11.7ms                      & 66.17                     & 156K                        \\ \hline
RA                            & 92,611M                      & 171,966M                     & 10.4ms                      & 12.0ms                       & 62.19                     & 163K                        \\ \hline
UDTA                          & 97,357M                      & 22,499M                      & 9.7ms                      & 2.6ms                      & 67.56                     & 865K                        \\ \hline
\end{tabular}%
}

\label{Table1}
\end{table}
\textbf{Experimental Environment and Settings}
We conducted experiments on a PC with RTX-3090 GPU, AMD Threadripper 1900x CPU, and 32GB of RAM. We implemented the models on Ubuntu Linux using PyTorch (1.7.1v) and Torchvision (0.8.2v).
The experimental settings and hyper-parameters are presented in \cref{tab:Experimental Settings}.

%For the hyperparameter, we used the pre-training epoch as 120, the learning rate as 0.01, Adam as the optimization function, batch size as 256, 0.001 as a learning rate and a cross entropy loss function. The other hyperparameters were  the same as those in the pretrained model.\par

\textbf{Metrics}
We measured classification performance by top-1 accuracy on the test data, the amount of computation by FLOPs (floating point operations), and time by msec. In each table, F.FLOPs and B.FLOPs denote the  computation for forward and backward propagation, and F.Time and B.Time denote the time for forward and backward propagation, respectively. Additionally, we present the number of parameters to train in each model.

%The performance of the model was measured using four components. Top-1 Accuracy (Acc.) is an index indicating the recognition performance of the model and is expressed as the ratio of the data that predicted the correct class to the entire dataset. FLOPs is the total sum of addition and multiplication operations (F.FLOPs, B.FLOPs) performed in the forward propagation and backpropagation learning process for one batch, and is a representative of the overall amount of computation. The learning time was measured in seconds (F.Time, B.Time) for forward propagation and backpropagation learning for one batch. Finally, the effectiveness of the parameters was evaluated by measuring the number of parameters used while training the model.

\begin{figure}[!h] 
\centering
\subfigure[Accuracy (blue), computation for forward propagation (red) and backpropagation (orange)]{
\includegraphics[width=0.6\linewidth]{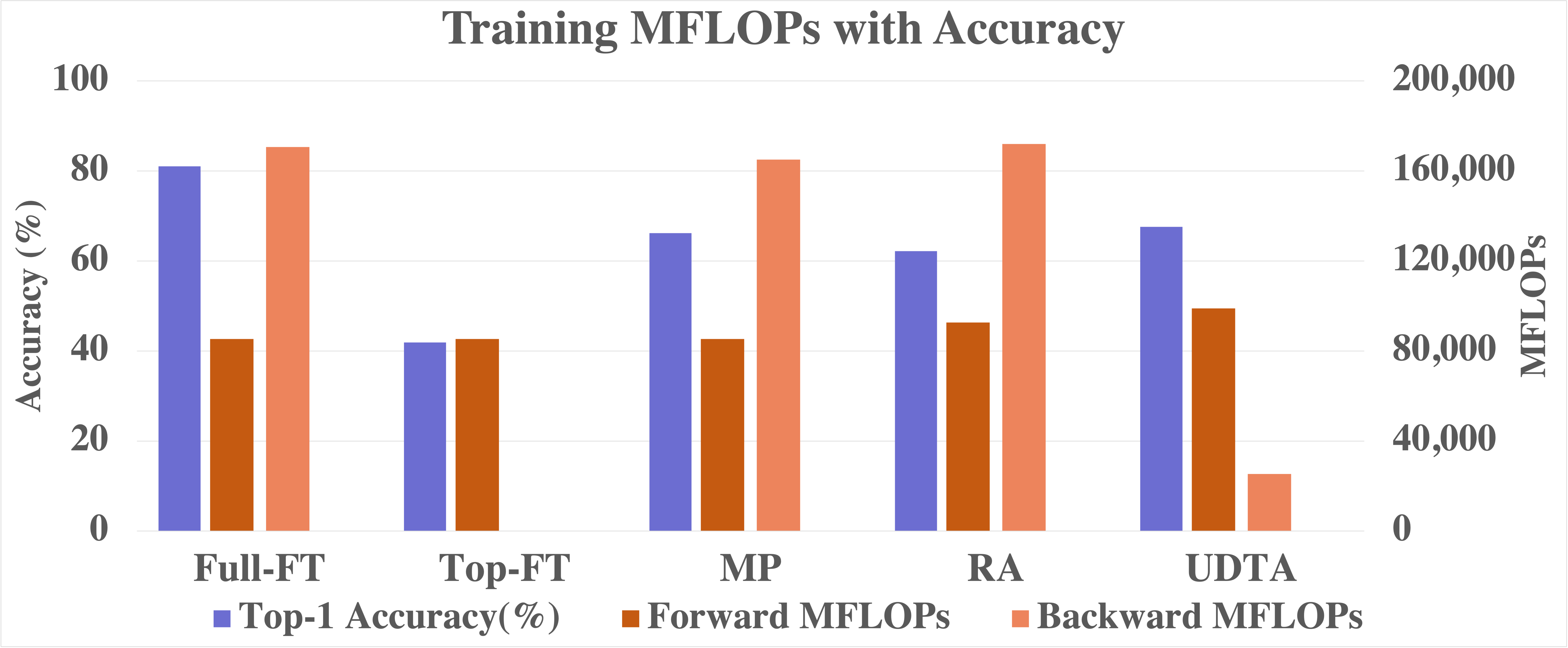}
}

\centering
\subfigure[Accuracy (blue), time for forward propagation (red) and backpropagation (orange)]{
\includegraphics[width=0.6\linewidth]{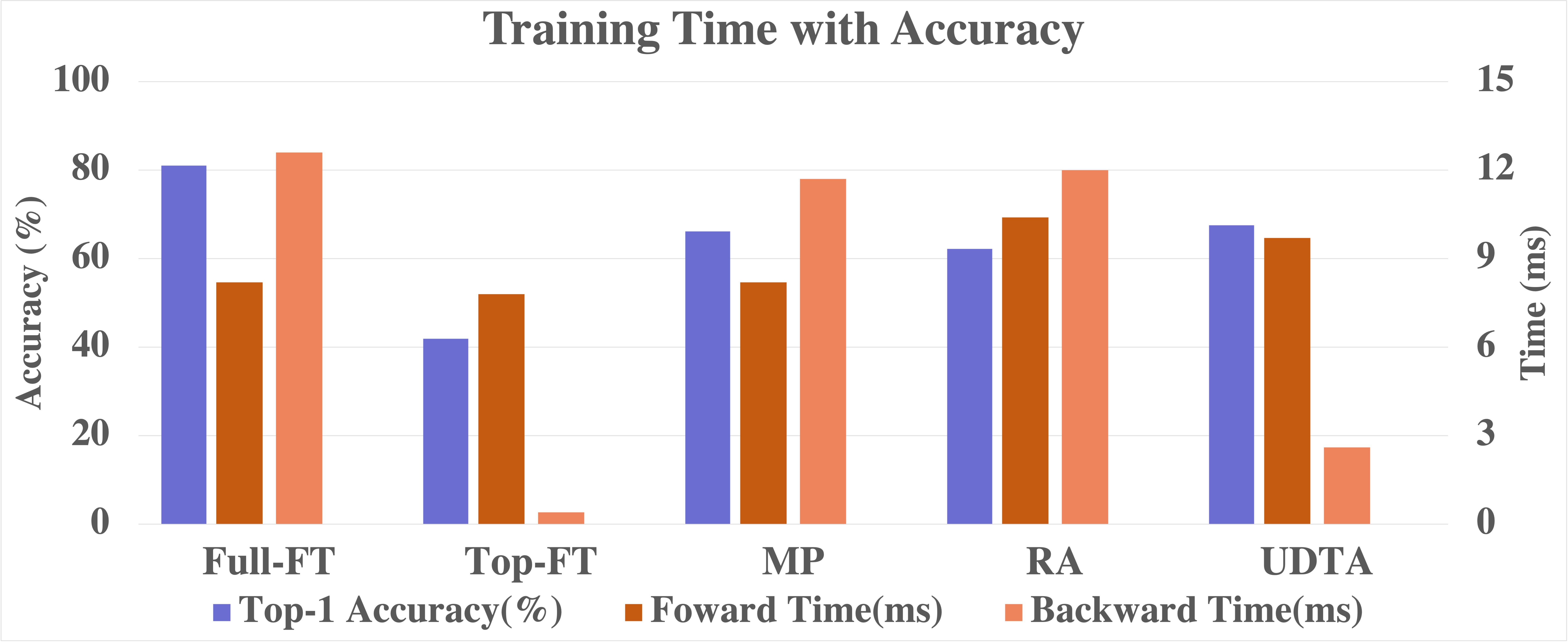}
}
\caption{Accuracy, computation, and time for forward/backward propagation of UDTA compared with other models} 
\label{Graphs}
\end{figure}

\subsection{Comparison with Other Adapter Networks}

In the first experiment, we compared UDTA with five other models on the FGVC-Aircraft dataset. The experimental results are presented in \cref{Table1} and Figure \ref{Graphs}, where `Scratch' denotes the backbone network only trained with the target datasets from scratch, `Full-FT' denotes the backbone pre-trained with ImageNet and then fully fine-tuned with the target dataset. `Top-FT' is the backbone network in which only the top layer was fine-tuned after pre-training. `MP', `RA', and `UDTA' denote model patch, residual adapter, and the proposed adapter, respectively.

Without pre-training, the backbone trained from scratch (`Scratch') was slow and exhibited a poor accuracy. The full fine-tuning of the pre-training backbone (`Full-FT') exhibited the best accuracy but required heavy computation for backpropagation. On the other hand, fine-tuning only the top layer of the pre-trained backbone (`Top-FT') was the fastest but exhibited the lowest accuracy. Applying model patches (`MP') and residual adapters (`RA') improved accuracy compared with the 'TOP-FT' model with learning a minimal number of parameters. However, they have little effect in reducing the computation and time for training. The proposed model, UDTA, exhibited higher accuracy than other adapters, while reducing the computation and time for backward propagation by 86.36\% from 165,096 MFLOPs to 22,499 MFLOPs, and 77.78\% from 11.7 msec to 2.6 msec compared with model patches. %The relative reduction rate in computation and time for backpropagation are 84.64\% and 73.50\%, respectively. 
In return, UDTA increased the computation and time for forward propagation by 15.88\% and 18.29\%, respectively.
%\cref{Table1} is the result of comparing the performance of the methods proposed in the previous study to the UDTA-applied model, and the graph in \cref{Graphs} shows the computation requirement (top) and the amount of change in learning time (bottom).\par
%With learning using UDTA, the computation requirement and time during forward propagation slightly increased by 15.92\% and 18.29\%, respectively, compared to the learning results for the entire model, but the computation requirement , and learning time consumed during backpropagation decreased significantly to 84.04\% and 75.4\%, respectively. Compared with the experimental results of the model patch and residual adapter, the amount of computation and training time decreased to a similar extent. The training performance decreased by approximately 12.63\% compared to the overall training result of the model, but improved performance compared to the model patch and residual adapter. Based on these experimental results, it was shown in the experiment that applying UDTA is more efficient than other existing methods with limited resources.\par

\begin{table}[t]
\caption{Computation, time, and accuracy according to the number of thin blocks B}

\centering
\resizebox{0.6\linewidth}{!}{%
\begin{tabular}{|c|c|c|c|c|c|c|}
\hline
B. & F.FLOPs  & B.FLOPs & F.Time & B.Time & Acc.  & Params \\ \hline
1  & 91,364M  & 12,148M & 8.2ms & 1.3ms & 65.01 & 634K   \\ \hline
3  & 97,357M  & 22,499M & 9.7ms & 2.6ms & 67.56 & 865K   \\ \hline
5  & 102,138M & 30,782M & 11.1ms & 3.7ms & 63.51 & 1002K  \\ \hline
7  & 113,233M & 51,870M & 11.7ms & 4.3ms & 60.24 & 1007K  \\ \hline
\end{tabular}%
}

\label{Table2}
\end{table}

\begin{table}[t]
\caption{Computation, time, and accuracy according to expansion factor $u$}
\centering
\resizebox{0.6\linewidth}{!}{%
\begin{tabular}{|c|c|c|c|c|c|c|}
\hline
$u$ & F.FLOPs  & B.FLOPs & F.Time & B.Time & Acc.  & Params \\ \hline
2 & 90,475M  & 8,570M  & 8.8ms & 2.1ms & 60.06 & 433K   \\ \hline
4 & 94,674M  & 16,967M & 9.3ms & 2.5ms & 65.22 & 705K   \\ \hline
6 & 97,357M  & 22,499M & 9.7ms & 2.6ms & 67.56 & 865K   \\ \hline
8 & 103,072M & 33,763M & 9.8ms & 3.3ms & 67.56 & 1249K  \\ \hline
\end{tabular}%
}
\label{Table3}
\end{table}

\begin{table}[!h]
\caption{Domain adaptation performance (Top-1 accuracy) on multiple datasets}
\centering
\resizebox{0.6\linewidth}{!}{%
\begin{tabular}{|c|r|r|r|r|r|}
\hline
        & \multicolumn{1}{c|}{Air} & \multicolumn{1}{c|}{Cars} & \multicolumn{1}{c|}{DTD} & \multicolumn{1}{c|}{Dogs} & \multicolumn{1}{c|}{Flow} \\ \hline
Scratch & 63.57\% & 72.80\% & 39.94\% & 55.65\% & 59.39\% \\ \hline
Full-FT & 81.03\% & 88.62\% & 69.14\% & 82.59\% & 95.65\% \\ \hline
TOP-FT  & 41.91\% & 50.20\% & 67.44\% & 80.17\% & 90.82\% \\ \hline
MP      & 66.17\% & 76.33\% & 68.61\% & 81.62\% & 93.78\% \\ \hline
RA      & 62.19\% & 74.94\% & 67.58\% & 77.87\% & 92.60\% \\ \hline
UDTA    & 67.56\% & 81.21\% & 69.89\% & 79.94\% & 93.54\% \\ \hline
\end{tabular}%
}

\label{Table6}
\end{table}

To find the optimal depth and width of UDTA, we measured computation, time, and accuracy on the FGVC-Aircraft dataset changing the number of thin blocks and the expansion factor $u$ in \cref{fig:architecture_of_UDTA}. The results are presented in \cref{Table2} and \ref{Table3}. As the width and depth of UDTA increased, the amount of computation and time increased. The accuracy increased with width but was not proportional to depth. We found the optimal size of UDTA, in terms of the trade-off between computation and accuracy, with depth three and expansion factor six.

%\cref{Table2} shows the experimental results obtained by changing the number of thin blocks of UDTA. The recognition performance of UDTA composed of three thin blocks was higher than that of UDTA composed of one thin block. However, beyond that, the recognition performance decreased as the number of thin blocks increased. \par
% In addition, in order to evaluate the effect of the change of the expansion coefficient on the performance in the process of extending the channel, an experiment was conducted to the effectiveness of the value of the expansion coefficient, and the experimental results are shown in \cref{Table3}. When the expansion coefficient is increased to 6, the amount of computation and learning time significantly increases, whereas increasing it to over 6 hardly seems to affect the prediction performance. 
% \par
% Because UDTA is able to flexibly change the depth and expansion coefficient of the model, it is possible to determine an appropriate trade-off between the amount of computation, learning time, and prediction performance according to the observed performance. In this study, based on the above experimental results, the UDTA structure of three thin blocks and an expansion factor of 6 that presented the best prediction performance was selected.\par

% \usepackage{graphicx}

We also evaluated UDTA and existing models on multiple fine-grained classification datasets.
\cref{Table6} measures the accuracy of the models. The `Full-FT' model exhibited the highest accuracy. The accuracy of UDTA was comparable or higher than the other adapter networks. \par
% \usepackage{graphicx}

\section{Conclusion}
We analyzed why conventional adapter networks fail to reduce the amount of computation for training and proposed a unidirectional thin adapter (UDTA) for efficient domain adaption of the DNN. UDTA improves the performance for the target task by providing auxiliary features to complement the features of the backbone network, which was trained from a general pre-training dataset. Unlike conventional adapter networks, UDTA is connected to the backbone network in one way, not requiring the gradient of the backbone network for training, which substantially reduces computation for backpropagation. In experiments on five small-size datasets for fine-grained classification, UDTA reduced the amount of computation and time for backpropagation by 86.36\% and 77.78\%, respectively, compared with model patches \cite{mudrakarta2018k}, while exhibiting comparable or even improved accuracy.

%Bibliography
\newpage

\bibliographystyle{unsrt}  
\bibliography{references}  
\newpage
\appendix
\onecolumn
\section{Training procedure}

The training of UDTA is composed of three steps: pre-training step, autoencoder training step, and domain adaptation step.

\begin{enumerate}
  \item Pre-train the backbone network using the source dataset which consists of a large amount of image data.
  \item Train the autoencoders using the source dataset. Each autoencoder is trained by the output features of the backbone network blocks which are connected to the autoencoders.
  \item Adapt the image recognition model(backbone network + encoder + UDTA) to a new domain using fine-grained classification dataset
  \begin{enumerate}[label=(\alph*)]
    \item Connect UDTA to the backbone network by the encoder.
    \item Train UDTA for domain adaptation while freezing the backbone network blocks. 
  \end{enumerate}
\end{enumerate}

% \begin{comment}
\begin{algorithm}[h]
   \caption{Pre-training algorithm}
   \label{alg:Pre-training algorithm}
\begin{algorithmic}
   \State \textbf{Source dataset}: $X_s = \{x_s^1, x_s^2, ... x_s^n\}$ where $n$ is the total number of the source data
   \State \textbf{Class labels of source dataset}: $Y_s = \{y_s^1, y_s^2, ... y_s^n\}$
   \State \textbf{Batch size}: $m$
   \State \textbf{Backbone network parameters}: $\theta_b$
   \State \textbf{Classification loss function}: $L$
   \State \textbf{Learning rate}: $\eta$
   \State
   \For{$i=1$ {\bfseries to} number of epochs}
       \For{$j=1$ to $n / m$}
       \State Sample $m$-size batch from $X_s$
       \State $k$ is the index of the data in the batch 
       \State Gradient $\hat{g} \gets \frac{1}{m} \nabla_{\theta_b} \Sigma_k L(f(x_s^k;\theta_b), y_s^k)$
       \State $\theta_b \gets \theta_b - \eta\hat{g}$ 
       
       \EndFor
   \EndFor

\end{algorithmic}
\end{algorithm}

\begin{algorithm}[h]
   \caption{Autoencoder training algorithm}
   \label{alg:example}
\begin{algorithmic}
  \State \textbf{Source dataset}: $X_s = \{x_s^1, x_s^2, ... x_s^n\}$ where $n$ is the total number of the source data
   \State \textbf{List of parameters of autoencoders}: $\Theta_{AE} = \{\theta_{ae}^1 , \theta_{ae}^2, ... \theta_{ae}^h\}$ where $h$ is the number of autoencoders  
   \State \textbf{Batch size}: $m$
   \State \textbf{Backbone network parameters}: $\theta_b$
   \State \textbf{Reconstruction loss function:} $L_{rec}$
   \State \textbf{Learning rate}: $\eta$
   \State
   \For{$i=1$ {\bfseries to} number of epochs}
       \For{$j=1$ to $n / m$}
           \State Sample $m$-size batch from $X_s$
           \State $k$ is the index of the data in the batch
           \For{$l=1$ to $h$}
               \State $o^l(x_s^k; \theta_b)$ is the $l$-th output feature of backbone network starting from the first block connected to the autoencoder
               \State $f(o^l(x_s^k; \theta_b); \theta_{ae}^l)$ is the output of $l$-th autoencoder
               \State Gradient $\hat{g}^l \gets \frac{1}{m}\nabla_{\theta_{ae}^l} \Sigma_k L_{rec}(f(o^l(x_s^k; \theta_b); \theta_{ae}^l), o^l(x_s^k; \theta_b))$
               \State $\theta_{ae}^l \gets \theta_{ae}^l - \eta\hat{g}^l$
           \EndFor
       \EndFor
   \EndFor

\end{algorithmic}
\end{algorithm}
\newpage
\begin{algorithm}[h]
   \caption{Domain adaptation algorithm}
   \label{alg:Pre-training algorithm}
\begin{algorithmic}
   \State \textbf{Target dataset}: $X_t = \{x_t^1, x_t^2, ... x_t^n\}$ where $n$ is the total number of the target data
   \State \textbf{Class labels of target dataset}: $Y_t = \{y_t^1, y_t^2, ... y_t^n\}$
   \State \textbf{Batch size}: $m$
   \State \textbf{Backbone network parameters}: $\theta_b$
   \State \textbf{Encoder parameters}: $\theta_e$
   \State \textbf{UDTA parameters}: $\theta_u$
   \State \textbf{Classification loss function}: $L$
   \State \textbf{Learning rate}: $\eta$
   \State
   \For{$i=1$ {\bfseries to} number of epochs}
       \For{$j=1$ to $n / m$}
           \State Sample $m$-size batch from $X_t$
           \State $k$ is the index of the data in the batch
           \State Gradient $\hat{g} \gets \frac{1}{m}\nabla_{\theta_u} \Sigma_k L(f(x_t^k;\theta_b, \theta_e, \theta_u), y_t^k)$
           \State $\theta_u \gets \theta_u - \eta\hat{g}$ 
       \EndFor
   \EndFor

\end{algorithmic}
\end{algorithm}
% \end{comment}
\newpage
\section{Specification of UDTA attached to MobileNetV2}
\begin{table}[h]
\caption{Structure of (a) MobileNetV2, (b) Encoders and (c) UDTA.}
\subtable[MobileNetV2]{

\centerline{\resizebox{0.64\textwidth}{!}{
        \begin{tabular}{c|c|c|c|c|c}
            \hline
            Input              & Operator    & Expansion  & Output & Repetition & Stride \\\hline
            $224^2$ X $3$      & Conv2d      & - & $112^2$ X $32$   & 1 & 2 \\
            $112^2$ X $32$     & IR block  & 1 & $112^2$ X $16$   & 1 & 1 \\
            $112^2$ X $16$     & IR block  & 6 & $56^2$ X $24$   & 2 & 2\\
            $56^2$ X $24$      & IR block  & 6 & $28^2$ X $32$   & 3 & 2 \\
            $28^2$ X $32$      & IR block  & 6 & $14^2$ X $64$   & 4 & 2 \\
            $14^2$ X $64$      & IR block  & 6 & $14^2$ X $96$   & 3 & 1 \\
            $14^2$ X $96$      & IR block  & 6 & $7^2$ X $160$  & 3 & 2 \\
            $7^2$ X $160$      & IR block  & 6 & $7^2$ X $320$  & 1 & 1 \\
            $7^2$ X $320$      & Conv2d 1x1  & - & $7^2$ X $1280$ & 1 & 1  \\
            $7^2$ X $1280$     & Avgpool 7x7 & - & $1^2$ X $1280$    & 1 & -   \\
        \end{tabular}%
}}}%
\par
\subtable[Encoders]{
    \centerline{\resizebox{0.4\textwidth}{!}{%
            \begin{tabular}{c|c|c|c}
                \hline
                Input              & Operator    & Output    & Stride \\
                \hline
                $7^2$ X $160$      & Conv2d 1x1  & $7^2$ x $80$  & 1 \\
                $7^2$ X $320$      & Conv2d 1x1  & $7^2$ x $160$  & 1 \\
                \end{tabular}%
}}}%
\par
\subtable[UDTA]{
    \centerline{\resizebox{0.49\textwidth}{!}{%
\begin{tabular}{c|c|c|c|c}
\hline
Input              & Operator    & Expansion & Output    & Stride \\
\hline
$14^2$ X $96$      & IR block  & 6 & $7^2$ x $80$  & 2 \\
$7^2$ X $80$      & IR block  & 6 & $7^2$ x $160$   & 1 \\
$7^2$ X $160$      & IR block  & 6 & $7^2$ x $320$ & 1 \\
$7^2$ X $320$      & Avgpool 7x7  & - & $1$x$1$ x $320$ & - \\
\end{tabular}
}}}%

\label{tab:specific sturcture}
\end{table}
In table (a), each line describes a sequence of one or more repeated operators. inverted residual block (IR block) consists of 1 x 1 convolution, 3 x 3 depth-wise convolution and 1 x 1 convolution layers. The depth-wise convolution in the first repetition of the sequence uses a stride represented in the tables and all others use a stride of 1 \cite{sandler2018mobilenetv2}. The encoder consists of a 1 x 1 convolution layer.
\newpage
\section{Experiments on training encoders separately}
We conducted an experiment on training UDTA and encoder during domain adaptation, comparing with solely training UDTA to scrutinize the effectiveness of training encoder separately. The separate training of the encoder achieves significantly increased accuracy with fewer parameters and less computation.
\begin{table}[h]
\caption{Computation, time, and accuracy of training only UDTA compared with training UDTA and encoder during domain adaptation. The separate training of autoencoder reduced the computation for backward propagation by 6.78\% from 24,135 MFLOPs to 22,499 MFLOPs and the number of parameters by 6.89\% from 929K to 865K. The accuracy increased from 65.43\% to 67.56\%}
\centering
\resizebox{0.7\linewidth}{!}{%
\begin{tabular}{|c|r|r|r|r|r|r|}
\hline
\multicolumn{1}{|c|}{Methods} & \multicolumn{1}{c|}{F.FLOPs} & \multicolumn{1}{c|}{B.FLOPs} & \multicolumn{1}{c|}{F.Time} & \multicolumn{1}{c|}{B.Time} & \multicolumn{1}{c|}{Acc.} & \multicolumn{1}{c|}{Params} \\ \hline
UDTA + Enc                            & 97,357M                      & 24,135M                     & 9.7ms              & 2.6ms                      & 65.43                     & 929K                        \\ \hline
UDTA                          & 97,357M                      & 22,499M                      & 9.7ms                      & 2.6ms                      & 67.56                     & 865K                        \\ \hline
\end{tabular}%
}

\label{tab:the effect of autoencoder}
\end{table}

\section{Sample datasets}
\begin{figure}[h] 
\centering
\subfigure[Oxford 102 Flowers image samples.]{
\includegraphics[width=0.65\linewidth]{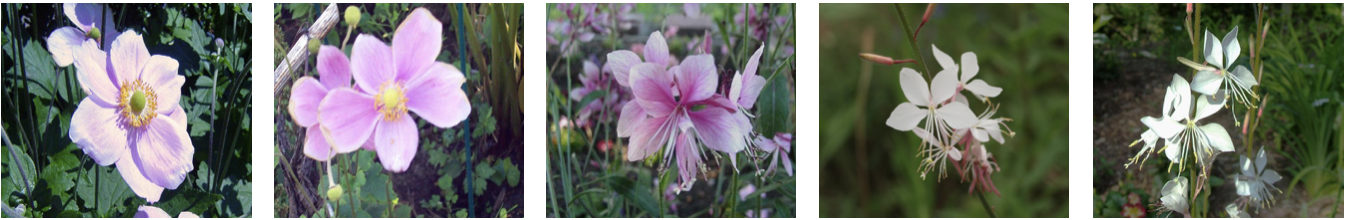}
}
\centering
\subfigure[DTD image samples.]{
\includegraphics[width=0.65\linewidth]{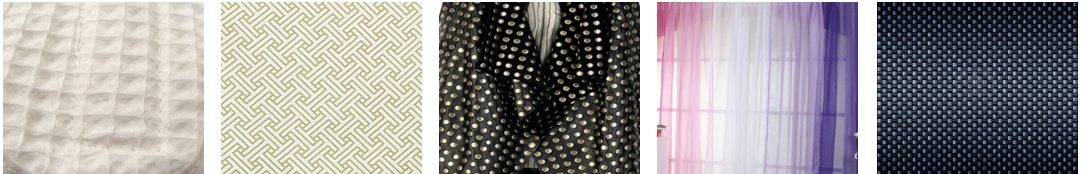}
}
\centering
\subfigure[Stanford Cars image samples.]{
\includegraphics[width=0.65\linewidth]{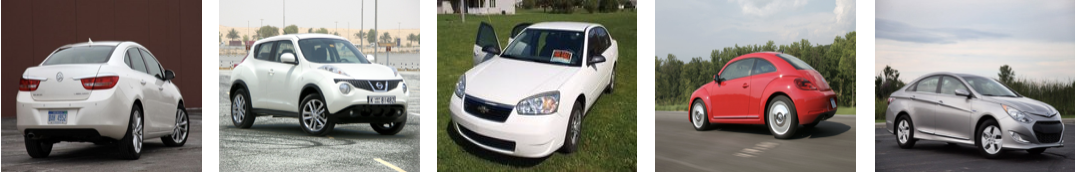}
}
\centering
\subfigure[Stanford Dogs image samples.]{
\includegraphics[width=0.65\linewidth]{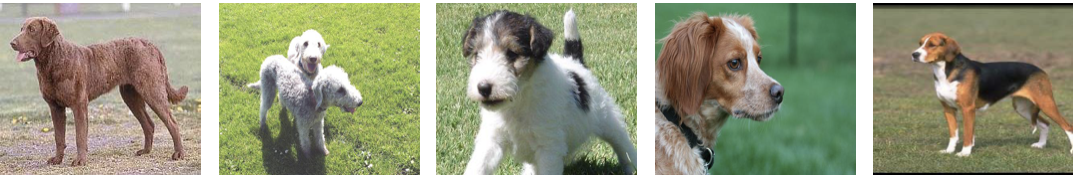}
}
\caption{Besides FGVC-Aircraft shown in \cref{fig:aircraft_sample}, (a) Oxford 102 Flowers, (b) DTD, (c) Stanford Cars, and (d) Stanford Dogs are used for domain adaptation experiments. They are fine-grained datasets consisting of a small number of images. Fine-grained classification datasets are composed of limited categories with barely discernible details which make them difficult to be distinguished. It is even more challenging for DNN to learn enough knowledge of those domains by a small number of images.}
\end{figure}

%\end{CJK}

\end{document}